\pdfoutput=1

\documentclass[11pt]{article}
\usepackage{acl}
\usepackage{courier}
\usepackage{cleveref}
\usepackage{graphicx}
\usepackage{array}
\usepackage{booktabs}
\usepackage{multirow}
\usepackage{hhline}
\usepackage{arydshln}
\usepackage{times}
\usepackage{latexsym}
\usepackage{hyperref}
\usepackage{bbding}

\usepackage[T1]{fontenc}

\usepackage[utf8]{inputenc}

\usepackage{microtype}
\usepackage{xcolor}
\usepackage{soul}

\newcommand{\entityannot}[2]{\textcolor{#1}{\textit{#2}}}
\newcommand{\eventannot}[2]{\textcolor{#1}{\textbf{#2}}}

\title{Contrastive Representation Learning for Cross-Document Coreference Resolution of Events and Entities}

\author{Benjamin Hsu \and Graham Horwood\\
AWS AI Labs \\
  \texttt{\{benhsu, ghorwood\}@amazon.com}
 }
  
\begin{document}
\maketitle
\begin{abstract}
Identifying related entities and events within and across documents is fundamental to natural language understanding. We present an approach to entity and event coreference resolution utilizing contrastive representation learning. Earlier state-of-the-art methods have formulated this problem as a binary classification problem and leveraged large transformers in a cross-encoder architecture to achieve their results. For large collections of documents and corresponding set of $n$ mentions, the necessity of performing $n^{2}$ transformer computations in these earlier approaches can be computationally intensive. We show that it is possible to reduce this burden by applying contrastive learning techniques that only require $n$ transformer computations at inference time. Our method achieves state-of-the-art results on a number of key metrics on the ECB+ corpus and is competitive on others.
\end{abstract}

\section{Introduction}

Coreference resolution is the fundamental NLP task of finding all mentions that refer to the same real world entity or event in text. It is an important step for higher level NLP tasks involving natural language understanding, such as text summarization \citep{azzam-etal-1999-using}, information extraction \citep{zelenko-etal-2004-coreference}, and question-answering \citep{vicedo-ferrandez-2000-importance}. Historically, coreference resolution of entities in the same text document -- within document (WD) coreference resolution -- has received the most attention, though more recently focus has moved toward cross-document (CD) coreference resolution.

CD coreference resolution has recently gained renewed interest for its application in multi-document analysis tasks. CD coreference resolution presents unique challenges not found in the WD context. Spans of text come from different documents without any inherent linear order, and there is no notion that antecedents for a given expression typically occur before the expression, as in a single document. Coreferent expressions also cannot be assumed to occur near one another. Furthermore, documents are also assumed to be authored independently and about different---though lexically similar---topics. For instance, the event described in the sentences from Topic 19 in Table \ref{tab:ecb_examples} below are not coreferential, despite their lexical similarity ("killed").

\begin{table*}[t!]
    \small
    \centering
    \begin{tabular}{lm{.4\textwidth}||m{.4\textwidth}}
    \hline
        & \begin{center} Subtopic 1 \end{center} & \begin{center} Subtopic 2 \end{center} \\
        \hline
        \multirow{5}{1.5cm}{Topic 19} &  Riots Erupt Following Death of \entityannot{orange}{Brooklyn} \entityannot{blue}{Teen} \eventannot{red}{Killed} By \entityannot{green}{Police} & INITIAL results from the post-mortem on a 15-year-old Greek \entityannot{cyan}{boy} \entityannot{cyan}{whose} \eventannot{magenta}{killing} by \entityannot{violet}{police} sparked five days of rioting show \entityannot{cyan}{Alexandros Grigoropoulos} \eventannot{magenta}{died} from a bullet ricochet. \\
        \cline{2-3}
        & Yesterday , the \entityannot{green}{police} explained that officers shot and \eventannot{red}{killed} a 16-year-old \entityannot{blue}{Kimani Gray} in \entityannot{orange}{Brooklyn} because \entityannot{blue}{he} allegedly pointed a gun at the cops. & Fresh riots were reported in Greece on Saturday December 13 2008 in protest at the \eventannot{magenta}{killing} by \entityannot{violet}{police} of a 15-year-old boy, \entityannot{cyan}{Alexandros Grigoropoulos}, eight days ago\\
        \hline
    \end{tabular}
    \caption{Examples of cross-document coreference clusters from topics 19 of the ECB+ corpus. Bold text indicate events and the same color indicates that they belong in the same coreference cluster. The addition of lexically similar second subtopic (riots in Greece over teenagers death vs riots in Brooklyn over teenagers death) adds an additional challenge to the ECB+ corpus.}
    \label{tab:ecb_examples}
\end{table*}

Another important aspect of CD coreference resolution is the potential scale of the problem. In certain applications, the number of documents can be large and ever growing. In particular, for applications that merge information from across documents, such as multi-document summarization \cite{falke-etal-2017-concept} or multi-hop question answering \cite{dhingra-etal-2018-neural}, the corpus in question can be both large and dynamically increasing in size. 

Past methods of CD coreference resolution have treated the problem as a binary classification task: given two pairs of mentions, classify them as referring to the same entity or not \cite{bejan-harabagiu-2010-unsupervised, TACL634, huang-etal-2019-improving, kenyon-dean-etal-2018-resolving}. In more recent works, contextual embeddings using a cross-encoder architecture have been leveraged to obtain state-of-the-art results \cite{yu2020paired, zeng-etal-2020-event, caciularu2021crossdocument} on the ECB+ corpus. Despite achieving state-of-the-art results on the benchmark dataset, a shortcoming of these approaches is the fact they use a transformer as a cross-encoder -- two sentences are passed to through the transformer network and a label is predicted. For $n$ mentions in a corpus, these approaches require $n^{2}$ comparisions at inference time. As \citet{reimers-gurevych-2019-sentence} noted when using BERT in a cross-encoder architecture, finding the most similar pair of sentences in a collection of $n = 10 000$ sentences  requires $n(n-1)/2 = 49\, 995\, 000$ inference computations, which they estimated to take 65 hours using a V100 GPU.

Others have sought to address the quadratic scaling of these methods. Recently, \citet{allaway2021sequential, cattan2021crossdocument} introduced methods that require $n$ transformer passes. In this work, we introduce a method using contrastive learning to generate mention representations that are useful for the coreference resolution problem. Previous attempts along these lines by \citet{kenyon-dean-etal-2018-resolving} introduced clustering-oriented regularization terms in the loss function. Our method improves on these earlier methods on the benchmark dataset, and achieves results competitive with the more expensive methods of \citet{yu2020paired, zeng-etal-2020-event, caciularu2021crossdocument}. We conduct extensive ablations of our model which we discuss in \textsection \ref{sec:ablations}. We discuss applications to domains outside of the ECB+ corpus in \textsection \ref{sec:TextRank}.

\section{Related Work}

Most recent work on CD coreference resolution has focused on the ECB+ corpus \cite{cybulska-vossen-2014-using}, which we also use in this work. The ECB+ corpus, which is an extension of the Event Coreference Bank (ECB), consists of documents from Google News clustered into topics and annotated for event coreference \cite{bejan-harabagiu-2010-unsupervised}. ECB+ increases the difficulty level of the original ECB dataset by adding a second set of documents for each topic (subtopic), discussing a different event of the same type (e.g. riots in Greece over teenagers death vs riots in Brooklyn over teenagers death; see Table \ref{tab:ecb_examples}) \cite{cybulska-vossen-2014-using}. While relatively small, the corpus is representative of the common cross-document coreference use cases across a restricted set of related documents (i.e. results from a search query). 

Most approaches to CD coreference resolution address the problem as a binary classification problem between all pairs of events and entities. Early works utilized hand engineered lexical features (e.g. head lemma, word embedding similarities, etc.) \cite{bejan-harabagiu-2010-unsupervised, TACL634}.  More recent works have relied on neural network methods, utilizing character-based embeddings \cite{huang-etal-2019-improving, kenyon-dean-etal-2018-resolving} or contextual embeddings \cite{yu2020paired, cattan2020streamlining, zeng-etal-2020-event, caciularu2021crossdocument, allaway2021sequential}. Recent approaches by \citet{yu2020paired} and \citet{caciularu2021crossdocument} leveraging RoBERTa and Longformer transformer models have set strong benchmarks. A drawback of these approaches is the necessity to consider all pairs of $n$ mentions in a corpus in a cross encoder architecture. Each unique pair of entities (separated by a special token) is passed through a transformer to generate a similarity score. This requires $n^{2}$ transformer computations.

This can be computationally expensive and several works have sought to address this.
\citet{allaway2021sequential} introduced a model that clusters mentions sequentially at inference time. They achieved competitive results using a BERT-base model and without using a hierarchical clustering algorithm to generate coreference chains. \citet{cattan2021crossdocument} adapted the model of \citet{lee-etal-2017-end} to the cross-document context. Specifically, they pruned document spans down to the gold mentions and encode each resulting pared document using a RoBERTa-large model. A pairwise (feed-forward network) scorer then generates a score for each pair of spans. They also considered an end-to-end system where they use their model to predict mention spans instead of using gold mentions. In this work, we consider gold mentions only as has been done in earlier works.

In this work, we introduce a method leveraging contrastive learning using a RoBERTa-large model as the base encoder. At inference time, our method requires $n$ passes of the transformer, like earlier methods by \citet{allaway2021sequential, cattan2021crossdocument}. Our method surpasses their methods on the benchmark ECB+ dataset and is competitive with more expensive cross-encoder approaches of \citet{yu2020paired, zeng-etal-2020-event, caciularu2021crossdocument}.

\section{Methodology}

\subsection{Dataset}

We follow earlier works and use the ECB+ corpus, which is an extension of the Event Coreference Bank (ECB), which was discussed in the previous section. Following earlier works by others \cite{yu2020paired,  cattan2020streamlining, caciularu2021crossdocument,allaway2021sequential}, we follow the setup of \citet{cybulska-vossen-2015-translating}, which was also used by others \cite{yu2020paired,  cattan2020streamlining, caciularu2021crossdocument,allaway2021sequential}. This setup uses a subset of the annotations which has been validated for correctness and allocates a larger portion of the dataset for training. In this setup, we use topics 1-35 as the train set, setting aside topics 2, 5, 12, 18, 21, 23, 34, 35 for hyperparameter tuning, and 36- 45 as the test set. To preprocess mentions, we utilized the reference implementation from \citet{cattan2020streamlining}. The distribution of the train, test, and development sets can be seen in Table \ref{table:ecb_stats}.

\begin{table}[ht!]
\centering
\begin{tabular}{lrrr}
\hline
 & \textbf{Train} & \textbf{Dev} & \textbf{Test} \\
\hline
\texttt{\#} Topics	& 25 & 8 & 10\\
\texttt{\#}  Documents	& 574 &	196	& 206 \\
\texttt{\#}  Event Mentions &	3808 & 1245 & 1780 \\
\texttt{\#}  Event Singletons & 1116 & 280	& 623 \\
\texttt{\#}  Event Clusters	& 1527 & 409 &	805 \\
\texttt{\#}  Entity Mentions	& 4758 &	1476 &	2055 \\
\texttt{\#}  Entity Singletons	& 814	& 205	& 412 \\
\texttt{\#}  Entity Clusters	& 1286 & 330 &608 \\
\hline
\end{tabular}
\caption{Statistics for the ECB+ corpus. We followed the setup of \cite{cybulska-vossen-2015-translating} and used topics 36-45 for our test set and topics 1-35 for training with topics 2, 5, 12, 18, 21, 23, 34, 35 set aside in the development set for hyperparameter tuning.}
\label{table:ecb_stats}
\end{table}

\subsection{Model}
\label{sec:model}
\begin{figure}[t]
\centering
    \includegraphics[width=0.75\linewidth]{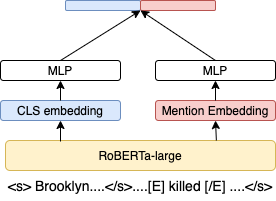}
    \caption{Our encoder takes as input the first two sentences from the document and concatenates it with the sentence containing the mention, taking care to annotate the mention location with tags $[E]$ and $[/E]$. }
    \label{fig:encoder}
\end{figure}

We propose a model to learn embeddings useful for clustering events and entities. Our model leverages a Siamese neural network \cite{10.5555/2987189.2987282} to fine-tune a RoBERTa-large encoder (see Figure \ref{fig:encoder}). 
We train and evaluate our model using gold mentions as opposed to predicted mentions in order to focus on the cross-document coreference resolution problem. At inference time, our model generates embeddings for the mentions which are then clustered using an agglomerative clustering algorithm as was done previously by \citet{barhom-etal-2019-revisiting, yu2020paired, cattan2020streamlining, caciularu2021crossdocument, zeng-etal-2020-event}. Below we discuss details of our methodology and training procedure.

\paragraph{Document Context}
 Following \citet{caciularu2021crossdocument}, we use the observation that other parts of the document provide valuable context to the mentions in question. We extract and encode the first two sentences from the document. This takes advantage of the fact that the articles are news articles and in many cases, much of the relevant information is summarized at the beginning of the document. In most cases, these two sentences are the headline and dateline for the article. In cases where the sentence in question is one of the first two sentences, we take the next sentence in the document. 
 
 

\paragraph{Contextual Embedding}

In addition to the document context, we also utilize the sentence that the mention appears in and annotate its location in the sentence using $[E]$ and $[/E]$ tokens. The two sequences are concatenated together using a [SEP] token (see Figure \ref{fig:encoder}). In total, we keep 128 word piece tokens and in cases where the combined input exceeds this, we remove tokens from the end of the context before removing tokens from the sentence containing the mention.

This combined sequence is encoded using a RoBERTa-large model \cite{liu2019roberta}, as shown in Figure \ref{fig:encoder}. We fine-tune all layers of the RoBERTa-large model. RoBERTa will produce a representation vector for each token of the input sequence. We then sum up element-wise the token-level representations of the mention and use this as the representation of the mention, $v_{e}$. Additionally, we utilize the first token of the sequence $v_{cls}$ as the embedding for the entire document context and mention embedding. Each of these contextual embeddings are passed separately through a multi-layer perceptron (MLP). We found that 1024 for the hidden layer dimension for both MLPs worked well in our experiments.

\begin{equation}
v^{\prime}_e = MLP_{1}(v_{e}); \quad
v^{\prime}_{cls} = MLP_{2}(v_{cls})
\end{equation}

The final representation for the mention $i$ and its context document is given by the concatenation of the two vectors output vectors, indicated by $[. ; .]$.

\begin{equation}
v _{i}= [v^{\prime}_{cls}; v^{\prime}_{e}]
\end{equation}

At inference time, our model takes in the mention and its context (both the head of the document and its sentence) and generates a 2048 dimensional embedding $v_{i}$. A clustering algorithm is applied to embeddings to generate coreference clusters. In order to compare our language model with earlier approaches, we follow earlier works and use an agglomerative clustering model. We use the implementation from scikit-learn\footnote{https://scikit-learn.org} and cluster mention representations using the cosine distance metric.
Representations within an average threshold distance $\tau$ are considered to be in the same cluster (i.e. coreferences).

\begin{table}
\centering
\resizebox{\columnwidth}{!}{%
\begin{tabular}{lrr}
\hline
 & \textbf{Events} & \textbf{Entities} \\
\hline
\texttt{\#} of Pairs	& 19000	& 27090	\\
\texttt{\#} of Positive	& 2085 &	4078 \\	
\texttt{\#} of Negatives	& 16915 &	23012 \\
\texttt{\#} of Same Subtopic &	13694 &	18847\\	
\texttt{\#} of Different Subtopic &	5306	& 8243\\
Fraction Positive &	0.11 & 0.15\\	
Fraction Same Subtopic & 0.72 & 0.70\\	
Median pos. similarity score & 0.62 & 0.59	\\
Median neg. similarity score & 0.80 & 0.77 \\
\hline
\end{tabular}}
\caption{Statistics for the contrastive pairs generated. Pairs of sentences were chosen from \textit{within} gold topics and were constructed by first computing the similarity between sequences. Negative samples were down-sampled by selecting samples whose similarity was greater than the median similarity among all possible sample pairs.}
\label{table:pair_dist}
\end{table}

\subsection{Training}

To train the model, we consider pairs of sentences -- positive samples are pairs of sentences where the mentions are coreferential while negative samples are pairs of sentences where the mentions are not coreferential. Pairs of sentences were chosen from \textit{within} gold topics and were constructed by first computing the similarity between sequences. This focuses our model to learn features to distinguish between the two closely related subtopics, one of the key aspects of the ECB+ corpus. We used SBERT \cite{reimers-gurevych-2019-sentence} to embed these sequences initially. Positive pairs were created from sequences that were least similar to one another and negative pairs were selected from the set of pairs most similar to one another, both within a particular subtopic and across subtopics (but still within the same topic). Finally, the negative samples were down-sampled by selecting samples whose similarity was greater than the median similarity among all possible positive sample pairs. The resulting distribution for the pairs can be seen in Table \ref{table:pair_dist}.

The model parameters were then trained using a Siamese network architecture \cite{ 1467314} where model weights are shared across both branches. 
For a given pair of sentences $p = (s_{1}, s_{2})$ and label $y = {1,0}$ where $y=1$ if the pairs are coreferences and $y=0$ otherwise, each pair of sentences is encoded using our model. The model was trained by minimizing the contrastive loss, $\ell$ \cite{1640964}, as implemented by \citet{reimers-gurevych-2019-sentence},

\begin{equation}
\ell = y * d(i,j)^{2} + (1-y) * max(0, m-d(i,j))^{2}
\end{equation}

For our purposes, $d(i,j) = 1-cos(v_{i}, v_{j})$ is the cosine distance, $m > 0$ is a margin, and $y$ is one if the pairs describe coreferent mentions and zero otherwise. Dissimilar pairs contribute to the loss function only if their distance are within $m$. The loss pushes the embeddings so that positive pairs are closer together in the embedding space and negative pairs are pushed to be more distant than the margin $m$.

\subsection{Hyperparameters}
\label{sec:Hyperparams}
In our experiments, we used the AdamW optimizer without warmup and found that a batch size of 16 worked well. We utilized Ray \cite{liaw2018tune} for hyperparmeter tuning and specifically the Bayesian optimization search algorithm from scikit-optimize.\footnote{https://github.com/scikit-optimize/scikit-optimize} We performed our experiments on a p3dn.24xlarge with 8 V100 Tensor Core GPUs and chose the dropout rate, learning rate, contrastive margin $m$ and clustering threshold $\tau$ to optimize the CoNLL F1 score on the development set gold topics. This was done to learn representations that address the lexical ambiguity in the ECB+ corpus topics. Resulting hyperparameters can be found in Table \ref{tab:hyperparams}.

\begin{table}[ht]
    \centering
    \begin{tabular}{lc|c}
        \hline
               &  Events & Entities\\
        \hline
        Epochs & 100 & 50 \\
        Learning rate & 2e-7 & 2e-7 \\
        Batch Size & 16 & 16 \\
        Contrastive margin, $m$ & 0.40 & 0.70 \\
        Clustering Threshold, $\tau$ & 0.2 & 0.2 \\
        \hline
    \end{tabular}
    \caption{Hyperparameters for our best performing models on events and entities.}
    \label{tab:hyperparams}
\end{table}

\section{Results and Discussion}

\begin{table*}[ht!]
\centering
\resizebox{\textwidth}{!}{%
    \begin{tabular}{lllccc|ccc|ccc|ccc|c}
\hline
 & & & \multicolumn{3}{c}{MUC}& \multicolumn{3}{c}{$B^{3}$} & \multicolumn{3}{c}{CEAF-e} & \multicolumn{3}{c}{LEA} & CoNLL \\
 \hhline{~~~|---|---|---|---|-}
 & &  &R  & P & F1 &   R   &     P       & F1  &  R &  P       & F1    &   R   &     P     & F1  &   F1 \\
\hline
\multirow{6}{0.75cm}{\centering Events} & \multirow{3}{1cm}{\centering Gold Topics} &  
Baseline & 72.9 &	72.4 &	72.7 &51.4 &	56.5 &	53.8 &	58.6 & 40.4	& 47.8	& 46.8	& 51.5 & 49.1 & 58.1
 \\
& & \citet{cattan2021realistic} & 80.1 & 76.3 & 78.1 & 63.4 &	54.1 & 58.4	& 56.3 & 44.2 &	49.5 & 59.7 & 49.6 &	54.2 & 62.0 \\
& & Ours & \textbf{87.8} &	\textbf{83.0} &	\textbf{85.3} &	\textbf{78.0} &	\textbf{71.4} & \textbf{74.5} &	\textbf{71.0} &	\textbf{57.4} &	\textbf{63.5} &	\textbf{75.6} &	\textbf{68.9} &	\textbf{72.1} &	\textbf{74.4}\\
\cline{2-16}
& \multirow{3}{1cm}{\centering Corpus} &
Baseline & 72.9 &	60.5 &	66.1 &	52 & 39.2 &	44.7 & 48.1 & 34.7 & 40.3 &	46.8 & 34.6 & 39.8 & 50.4 \\
& & \citet{cattan2020streamlining} & 79.9 & 74.8 & 77.2 & 62.2 &	48.9 & 54.8 & 53.3 & 42.3 &	47.2 &	58.4 & 44.4 &	50.5 &	59.7 \\
& & Ours & \textbf{86.4} & \textbf{74.9} & \textbf{80.2} & \textbf{76.2} & \textbf{51.7} & \textbf{61.6} & \textbf{59.7} &	\textbf{48.7} &	\textbf{53.6} &	\textbf{73.4} &	\textbf{48.7} &	\textbf{58.6} &	\textbf{65.2} \\
\hline
\hline
\multirow{6}{1cm}{\centering Entities} & \multirow{3}{1cm}{\centering Gold Topics} &  
Baseline & 61.6 & 85.9 & 83.2 & 31.8 & 80.2 &	45.5 & 53.7 & 33.8 & 41.5 &	28 & 76.9 &	41 & 52.9 \\
& & \citet{cattan2020streamlining} & \textbf{85.8} &	79.3 &	82.4 & 64.3 & 60 & 62.1 & 58.6 & 45.9 &	51.5 &	60.9 & 56.8 &	58.8 &	65.3 \\
& & Ours & 84.5 & \textbf{90.1} & \textbf{87.2} &	\textbf{72.2} &	\textbf{80.5} &	\textbf{76.2} &	\textbf{73.1} &	\textbf{59.0} & \textbf{65.3} & \textbf{69.7} & 78.5 & \textbf{73.8} &	\textbf{76.2}\\
\cline{2-16}
& \multirow{3}{1cm}{\centering Corpus} &
Baseline & 61.9 &	77.5 & 68.8 & 32.5 & 67.7 &	43.9 & 50.1 & 33.2 & 39.9 &	28.1 &	63.8 & 39 & 50.9 \\
& & \citet{cattan2020streamlining} & \textbf{85.7} & 79.3 & 82.4 & 63.7 &	60 & 61.8 &	58.1 & 45 & 50.7 &	60.3 & 56.8 &	58.5 &	65 \\
& & Ours & 83.9	& \textbf{86.6} & \textbf{85.2} &	\textbf{71.4} &	\textbf{75.6} &	\textbf{73.4} & \textbf{69.2} & \textbf{55.1} & \textbf{61.4} &	\textbf{68.5} &	73.1 &	\textbf{70.7} &	\textbf{73.3}\\
\hline
\end{tabular}}

\caption{\label{tab:wo_singletons} Combined within- and cross-document coreference scores for entities and events \textit{without} singletons, using gold mentions. Gold topics use the ECB+ topics as the initial document pre-clustering while corpus level results do not use any document pre-clustering. \textbf{Bold} values indicate best overall for a particular data subset.}
\end{table*}

We evaluate our model using four different measures as is common in earlier works. Specifically, we evaluated our model performance using MUC \cite{10.3115/1072399.1072405}, $B^{3}$ \cite{Bagga1998AlgorithmsFS}, CEAF-e \cite{luo-2005-coreference}, and LEA \cite{moosavi-strube-2016-coreference} metrics. We also evaluate our model using the CoNLL F1, the average of the MUC, $B^{3}$, and CEAF-e F1 scores. As a baseline, we also show results from a lemma model that takes each span in question and utilizes spaCy\footnote{https://spacy.io/} to lemmatize each token. Mentions are clustered based on whether their lemmatized tokens are exact matches or not.

Evaluations on ECB+ test corpus are not without controversy, and we discuss these subtleties in detail below. For the reader familiar with these issues, our main results are discussed in \textsection \ref{sec:gold_corpus} and \textsection \ref{sec:pred_topics}. We also conduct an ablation study with results in \textsection \ref{sec:ablations}.

\subsection{Evaluation Settings}

Many earlier methods leveraged an initial document clustering \cite{yu2020paired, zeng-etal-2020-event, caciularu2021crossdocument, allaway2021sequential}.  As observed by \citet{barhom-etal-2019-revisiting, upadhyay-etal-2016-revisiting}, clustering the documents as a preprocessing step and performing pairwise classification on mentions within each cluster provides a strong baseline. \citet{barhom-etal-2019-revisiting} introduced a K-Means algorithm to cluster documents using TF-IDF scores of the unigrams, bigrams and trigrams, where K is chosen by utilizing the Silhouette coefficient method \cite{ROUSSEEUW198753}. Models are then applied to mentions within each cluster.

However, this approach has come under criticism \cite{cremisini-finlayson-2020-new, cattan2021crossdocument, cattan2021realistic}. Detractors note that, because of the high lexical similarity between documents within the same subtopic, pre-clustering methods are able to produce near perfectly predicted subtopics, especially in the ECB+ corpus, where only a few coreference links are found across different subtopics. Document clustering is not expected to perform as well in realistic settings where coreferent mentions can spread over multiple topics \cite{cattan2021crossdocument}. More importantly, this bypasses the intention behind the inclusion of subtopics in ECB+ and avoids challenging the coreference models on lexical ambiguity \cite{cybulska-vossen-2014-using}.  

In our view, evaluation utilizing the original topic clusters ("gold" topics) is more in line with the original intent of \citet{cybulska-vossen-2014-using} and more indicative of realistic settings \cite{cattan2021realistic}. We discuss results (1) using ECB+ topics ("gold topics" henceforth) as the initial document clustering and (2) using no initial document clustering ("corpus level" henceforth) in section \textsection \ref{sec:gold_corpus}. We find that our methodology improves on earlier methods (Tables \ref{tab:wo_singletons} and \ref{tab:w_singletons}). Finally, because a majority of earlier works evaluate their models using predicted topics, we discuss our model performance under this setting in \textsection \ref{sec:pred_topics}. We report results from a single run.

\subsection{Gold Topics and Corpus Level} \label{sec:gold_corpus}

We evaluate our models using the ECB+ topics, in line with the intent of \citet{cybulska-vossen-2014-using} and earlier works by \citet{cattan2021crossdocument, cattan2021realistic}. According to those authors, this setting was designed to approximate an unclustered stream of news articles.

Additionally, as noted by \citet{cattan2020streamlining, cattan2021crossdocument}, the presence of singletons biases the results towards models that perform well on detecting all the mentions instead of predicting coreference clusters. Furthermore, in using gold mentions in the evaluation (like we do here), including singletons artificially inflates performance metrics 
\cite{cattan2021crossdocument}. We present our results without singletons (Table \ref{tab:wo_singletons}) using the reference implementation of \citet{moosavi-strube-2016-coreference}. In Appendix \ref{appendix:metrics}, we give results \textit{with singletons} in Table \ref{tab:w_singletons}.

On the gold topic and corpus level subsets, our model performs well. In all cases, we surpass the current state-of-the-art model on the CoNLL F1 metric for both event and entity coreference resolution by large margins without singletons (see Table \ref{tab:wo_singletons})). We suspect this improvement to be a feature of contrastive learning and methodology we used to choose pairs -- coreferential mentions are pushed closer together in the embedding space while mentions that are not coreferences are pushed further apart. We do observe a larger drop in performance in going from gold topics to the corpus level subsets. This is due to the choice in contrastive pairs, where negative examples come from the same gold topic.



Aside from improved performance, our methodology differs in some key aspects to the recent works by \citet{cattan2021crossdocument, cattan2021realistic, cattan2020streamlining}. Their methodology also leverages a RoBERTa-large model to embed documents, but breaks long documents into 512 word piece token chunks. The authors used as their feature vector for a span in question: the sum of the span embeddings, the embeddings for the span beginning and end, and a vector encoding the span length as their feature vector, which they feed into a pairwise classifier to generate pairwise scores. We on the other hand use the sentences containing the span in question and additional context sentences from the document, keeping a total of 128 word piece tokens. This additional context from the document, despite keeping fewer tokens, accounts for much of the performance gain. This is discussed in further detail in \textsection \ref{sec:ablations}.


\subsection{Predicted Topic Clusters} \label{sec:pred_topics}

\begin{table*}[ht!]
\centering
\resizebox{\textwidth}{!}{
 \begin{tabular}{lp{1cm}p{1cm}p{1cm}p{1cm}p{2.5cm}p{4cm}cccc}
\hline
 &Scaling&Adapt.&Fine-tuned&SRL&Encoder&System&MUC F1&$B^{3}$ F1&CEAF-$e$ F1&CoNLL F1\\
 \hline
 \multirow{11}{1cm}{Events} &\multirow{4}{1cm}{$n^{2}$}&&&&&Baseline & 76.7 & 77.5 & 73.2 & 75.7 \\
&&&\Checkmark&\Checkmark&BERT-large&\citet{zeng-etal-2020-event} & 87.5 & 83.2 & \textbf{82.3} &  84.3 \\
 &&&\Checkmark&\Checkmark&RoBERTa-large&\citet{yu2020paired} & 86.6 & 85.4 &81.3  &  84.4 \\
 &&\Checkmark&\Checkmark&&Longformer&\citet{caciularu2021crossdocument} & \textbf{88.1} & \textbf{86.4}  & 82.2 & \textbf{85.6} \\
 \cdashline{2-11}
 &\multirow{7}{1cm}{$n$}&&&&RoBERTa-large&\citet{cattan2021crossdocument} & 83.5 & 82.4  & 77.0 &  81.0 \\
 &&\Checkmark&&\Checkmark&BERT-base&\citet{allaway2021sequential} &82.2 & 81.1& 79.1& 80.8 \\
 &&&\Checkmark&&&Ours &\\
 &&&&&& -- RoBERTa-large & \underline{85.6} & \underline{84.8} & \underline{79.6} & \underline{83.3} \\
 &&&&&& -- RoBERTa-base & 84.0 & 82.4 & 79.0 & 81.8\\
 &&&&&& -- BERT-large & 82.8 & 82.3 &77.9 & 81.0 \\
 &&&&&& -- BERT-base & 79.8 & 79.4 & 74.4 & 77.9 \\
 \hline
 \multirow{9}{1cm}{Entities} &\multirow{2}{1cm}{$n^{2}$}&&&&&Baseline & 70.7 & 61.7& 56.9 & 63.1 \\
  &&\Checkmark&\Checkmark&&Longformer&\citet{caciularu2021crossdocument}& \textbf{89.9} & \textbf{82.1} & \textbf{76.8} & \textbf{82.9} \\
 \cdashline{2-11}
 &\multirow{6}{1cm}{$n$}&&&&RoBERTa-large& \citet{cattan2021crossdocument} &83.6 & 72.7 & 63.1 & 73.1 \\
&&\Checkmark&&\Checkmark&BERT-base&\citet{allaway2021sequential} &84.3  & 72.4 & 69.2 & 75.3 \\
 &&&\Checkmark&&Ours &\\
 &&&&&& -- RoBERTa-large & \underline{87.1} & \underline{80.3} & \underline{73.1} & \underline{80.2} \\
 &&&&&& -- RoBERTa-base & 83.6 & 74.1 & 68.5 & 75.4 \\
 &&&&&& -- BERT-large & 80.8 & 71.4 &66.2  & 72.8 \\
 &&&&&& -- BERT-base & 78.2 & 68.9 & 62.7 &  69.9 \\
 \hline
 \end{tabular}}
 \caption{\label{tab:subtopic_f1} A comparison of methods utilizing contextual embedding models and their performance on the ECB+ test corpus using \textit{predicted} topic clusters of \citet{barhom-etal-2019-revisiting}. We have indicated the scaling at inference time (in terms of transformer computations) above. We have also indicated whether systems utilized adaptive pre-training (Adapt.), fine-tuned encoders (Fine-tuned), or utilized a semantic role labelling model (SRL). To better compare to earlier works, we have included results from using different encoders in our model and indicated which encoders were used in earlier works. Finally, \citet{allaway2021sequential} used sequential clustering algorithm whereas ours and \citet{cattan2020streamlining} utilized an agglomerative clustering algorithm. \textbf{Bold} indicates best overall. \underline{Underlined} results indicate our best overall.}
\end{table*}

We compare our model against the majority of earlier works that used predicted topic clusters and gold mentions (see Table \ref{tab:subtopic_f1} and Appendix \ref{appendix:metrics} Table \ref{tab:subtopic} for more complete results). We used the reference implementation by \citet{pradhan-etal-2014-scoring} to score our models \textit{with singletons}. Our model is competitive with earlier approaches \cite{yu2020paired, zeng-etal-2020-event, caciularu2021crossdocument}, despite using significantly fewer resources at inference time -- $n$ transformer computations at inference time as oppose to $n^{2}$ transformer computations. 
We also note that in contrast to our approach, \citet{caciularu2021crossdocument} used a total of 600 tokens from each document (most documents are within 512 tokens) whereas we only use 128 tokens. Models by \citet{yu2020paired, zeng-etal-2020-event} employ a BERT based semantic role labelling (SRL) model. On average, our model lags their models by approximately 1.1 CoNLL F1 points, however, we note that \citet{yu2020paired} find that the SRL tagging accounted for roughly 0.4 CoNLL F1 points.

When comparing to other models that are linear in transformer computations, our model does well. Compared to the work by \citet{allaway2021sequential}, our model surpasses their results by 3.7 CoNLL F1 points on average. We note however, that their model used a BERT-base model and that they also introduced a novel sequential clustering approach. Our methodology used the larger RoBERTa-large model, and we utilized an agglomerative clustering algorithm as in previous works. 


Finally, in contrast to earlier works, we note that our model performs equally well when using predicted clusters and ECB+ gold topics. In fact, our model does better (by 0.9 CoNLL F1 points) on entities when going to gold topics, and achieves the same performance on events using gold topics. This is related to how we selected our contrastive pairs -- negative and positive pairs were selected from within each topic and so our model focused on the lexical ambiguity in the ECB+ corpus. 



\subsection{Training and Inference Time}
\label{sec:TrainingTime}
Our model is larger than earlier models by \citet{cattan2021crossdocument, cattan2021realistic, allaway2021sequential}. On a single V100 Tensor Core GPU with 32 GB of RAM, training took approximately two days. This is comparable to reported times for the cross-encoder model (using Longformer) by \cite{caciularu2021crossdocument}. We note that contrastive learning methods have been found to converge slowly \cite{NIPS2016_6b180037}. At inference time, however, our model takes approximately 15 seconds to evaluate on the ECB+ test set of events (using gold mentions and with singletons included). As a point of comparison, we ran the model of \citet{cattan2021crossdocument, cattan2021realistic} 
which likewise uses RoBERTa-large and is linear in transformer computations. We found that their model takes approximately 60 seconds under similar settings. In \textsection\ref{sec:ablations} we discuss experiments with smaller models.

\subsection{Ablations}
\label{sec:ablations}

\begin{table}[t]
\centering
\resizebox{\columnwidth}{!}{
 \begin{tabular}{llcc|cc}
\hline
 & & \multicolumn{2}{c}{\centering Entities} & \multicolumn{2}{c}{\centering Events} \\
 & & F1 & $\Delta$ & F1 & $\Delta$ \\
 \hline
 \multirow{5}{1cm}{\centering Pred. Topics} & Our Model & 80.2 & & 83.3 & \\
 &-- \texttt{CLS} representation & 77.8 & -2.4 & 82.3 & -1.0\\
 &-- mention representation & 77.8 & -2.4 & 82.0 & -1.3\\
 &-- no document context & 74.2 & -6.0 & 80.8 & -2.5 \\
 \hline
 \multirow{5}{1cm}{\centering Gold Topics} & Our Model  & 81.1 &  & 83.3 &\\
 &-- \texttt{CLS} representation & 79.0 & -2.1 & 81.3 & -2.0 \\
 &- mention representation & 78.9 & -2.2 & 79.6 & -3.7\\
 &-- no document context & 75.1 & -6.0 & 77.1 & -6.2\\
 \hline
 \multirow{5}{1cm}{\centering Corpus} & Our Model  & 78.7 &  & 75.9 &\\
 &-- \texttt{CLS} representation & 77.3 & -1.4 & 74.9 & -1.0\\
 &-- mention representation & 75.7 & -3.0 & 72.0 & -3.9\\
 &-- no document context & 72.8 & -6.0 & 70.5 & -5.4\\
 \hline
 \end{tabular}}
 \caption{\label{tab:ablations} Ablation results (CoNLL F1) on the ECB+ test set \textit{with singletons}.}
\end{table}

We ablate several parts of our model using the headlines heuristic and examine the importance of the underlying language model, the token representations, and the document context.

\paragraph{Language Model}
We examined the effect different representations have on overall performance by ablating the language model used. We found that the larger and richer representations of the RoBERTa-large model performed better generically. We gained on average 5 CoNLL F1 points in using RoBERTa-large versus BERT-large. We gained on average 7.2 CoNLL F1 points versus the smaller BERT-base model. Details can be found in Table \ref{tab:subtopic_f1}.


\paragraph{Token Representation}
To assess the effect of including the \texttt{CLS} token embedding in the final representations, we trained our model without using its representation, but keeping the mention representation. We find that the \texttt{CLS} representation accounts for roughly 1.3 CoNLL F1 points on average while the mention representation accounts for roughly 2.8 CoNLL F1 points on average (see Table \ref{tab:ablations} for details). We also examined our model without explicitly using the mention representation, but still tagging the span with $[E],[/E]$ tokens. For our model, we find that the mention representation was a more important factor when considering events. We speculate that tagging the mention location with $[E],[/E]$ tokens allows the transformer to attend to the mention. For events, which have a more complicated structure (e.g. arguments) this likely has a more important effect. 

\paragraph{Document Context}
Finally, an important component of our model was including the first two sentences of each document in the spirit of \citet{caciularu2021crossdocument}. For the ECB+ corpus, which is comprised of news articles, much contextual information is contained in the first two sentences of the document. We see that the document context contributes on average 5.4 CoNLL F1 points (see Table \ref{tab:ablations} for details). This is in line with our expectations for new articles and earlier observations by \citet{caciularu2021crossdocument}. We suspect the importance of this feature is due to a property of the ECB+ corpus that has been highlighted by others -- namely, the documents form fairly distinct clusters in themselves and so simple document embeddings are able to recover subtopics easily \cite{cattan2021crossdocument, cattan2021realistic, cremisini-finlayson-2020-new}. Note for instance that our model without using document context is competitive (compare with Table \ref{tab:subtopic_f1}) when using predicted topics for pre-clustering. Recently, \citet{eirew-etal-2021-wec} sought to address this issue by creating a the Wikipedia Events Coreference (WEC) dataset. Applying our model to the WEC dataset, we found that our results surpass their benchmark by large margins (CoNLL F1 of 89.3 versus 62.3 \cite{eirew-etal-2021-wec}) . We plan to discuss these results in further detail in future work. 

\subsection{TextRank}
\label{sec:TextRank}
A limitation of the current work is its specificity to formal text (i.e. news articles, Wikipedia articles). Given the importance of the headlines to our model, we also conducted experiments using the TextRank algorithm \cite{mihalcea-tarau-2004-textrank} to extract sentences that best summarize the content of the article instead of using the first two. We expect this method to be more applicable to less formal settings. We embedded each sentence in the document using SBERT and select the top two. On average we found that the headlines heuristic provided a 4.6 and 3.7 CoNLL F1 gain on on event and entity coreference resolution respectively (with singletons) over the TextRank extracted contexts (for detailed metrics see Table \ref{tab:subtopic} in Appendix \ref{appendix:metrics}). This is expected in the ECB+ context as the TextRank algorithm selects noisier sentences as compared to article headlines.

\section{Conclusions}

In this paper, we proposed a new model for within- and cross-document coreference resolution. We demonstrated that contrastive learning approaches are effective at learning representations for coreference resolution. We evaluated our model on gold topics and at the corpus level of the ECB+ corpus---with and without singleton mentions---and found that our approach surpasses current state-of-the-art methods by large margins. We also evaluated our models with an initial document clustering method and found that our model was competitive with earlier works. We presented extensive ablations of our model and discussed limitations of our work including model size, training time, application to formal text domains (i.e. news articles and Wikipedia), and use of agglomerative clustering to generate final coreference clusters. Interesting directions for future work would be testing the TextRank algorithm in less formal contexts (i.e. beyond news articles and Wikipedia articles), investigating higher-order tuples (e.g. triplets) to speed up model convergence, and extending our work to predicted mentions as opposed to gold mentions as has been done by others \cite{cattan2021crossdocument, cattan2021realistic}.



\section*{Acknowledgements}

The authors thank the anonymous reviewers for their advice and comments.

\section*{Ethical Considerations}

In this work, we used the ECB+ corpus \cite{cybulska-vossen-2014-using} which consists of news articles from the open domain. Our use was consistent with the intended use of the dataset. Our model does not contain any intentional biases. As discussed in \textsection \ref{sec:Hyperparams}, \textsection \ref{sec:TrainingTime}, we ran our experiments on a single p3dn.24xlarge with 8 V100 32GB GPUs. Model training and inference was relatively short and does not present ethical issues.
\bibliography{anthology,custom}
\bibliographystyle{acl_natbib}

\appendix

\section{Detailed Metrics}
\label{appendix:metrics}
Below we give detailed metrics \textit{with singletons}.
\begin{table*}[t!]
\centering
\resizebox{\textwidth}{!}{%
    \begin{tabular}{llccc|ccc|ccc|ccc|c}
\hline
\hline
 & & \multicolumn{3}{c}{MUC} & \multicolumn{3}{c}{$B^{3}$} & \multicolumn{3}{c}{CEAF-e} & \multicolumn{3}{c}{LEA} & CoNLL \\
  \hhline{~~|---|---|---|---|-}
 & &  R  & P & F1 &   R   &     P       & F1  &  R &  P       & F1    &   R   &     P     & F1  &   F1 \\
\hline
\multirow{12}{1cm}{\centering Events} & Baseline & 72.5 &	81.1 & 76.6 & 69.6 & 87.4 &	77.5 &	77.9 &	69 & 73.2 &	55.63 & 72.9 &	63.1 & 75.7 \\

& \citet{zeng-etal-2020-event} & 85.6 & \textbf{89.3} & 87.5 & 77.6 & 89.7 & 83.2 & 84.5 & 80.1 & \textbf{82.3} & - & - & - & 84.3 \\
& \citet{yu2020paired} & \textbf{88.1} & 85.1 &	86.6 & 86.1 & 84.7 & 85.4 &	79.6 & \textbf{83.1} & 81.3 & - & - & - & 84.4 \\
& \citet{caciularu2021crossdocument}& 87.1 &  89.2 & \textbf{88.1} & 84.9 & \textbf{87.9} &	\textbf{86.4} &	\textbf{83.3} & 81.2 & 82.2 & \textbf{76.7} &	\textbf{77.2} & \textbf{76.9} & \textbf{85.6} \\
\cdashline{2-15}
& \citet{cattan2021crossdocument} & 85.1 & 81.9 & 83.5 & 82.1 &	82.7 &	82.4 & 75.2 & 78.9 & 77 & 68.8 & 72 & 70.4 & 81 \\
& \citet{allaway2021sequential} & 81.7	& 82.8 & 82.2 &	80.8 & 81.5 &	81.1 & 79.8 & 78.4 & 79.1 &	- & - &	- & 80.8 \\
& Ours &&&&&&&&&&&&  \\
 & -- RoBERTa-large &  87.9 & 83.4 & 85.6 & \textbf{86.2} & 83.4 &	84.8  &	76.9 & 82.4 & 79.6 & 74.1 & 74.2 &	74.1 & 83.3  \\
 & -- RoBERTa-base &  83.6& 84.5&  84.0& 78.9&	 86.1 &	 82.4 &	 79.5 &	 78.5 &	 79.0 &	 67.1 &	 75.8 &	 71.2 &	81.8 \\
 & -- BERT-large &  82.9 &	 82.7 &	 82.8 &	 81.3&	 83.4 &	 82.3&	 77.8 &	 78.0& 	 77.9 &	 68.9 &	 72.5	& 70.6 &81.0\\
 & -- BERT-base &  80.3 &	 79.3 &	 79.8 &  78.0 &	 80.9 &	 79.4 &	 73.8 &	 75.0 &	 74.4 &	 63.4 &	 68.8 &	 66.0&	77.9 \\
 &   -- RoBERTa-large + TextRank  & 80.0 & 83.6 & 81.8 & 76.9 &	 86.4 &	 81.4 & 78.6 & 74.7 & 76.6 & 64.1 & 74.3 & 68.8 & 79.9 \\
\hline
\multirow{10}{1cm}{\centering Entities} & Baseline & 58.7 & 88.6 & 70.7 & 46.2 & 93.1	& 61.7 & 79.7 &	44.2 & 56.9 & 35.6 & 68.2 & 46.8 & 63.1 \\
& \citet{caciularu2021crossdocument}& \textbf{88.1} & \textbf{91.8} & \textbf{89.9} & \textbf{82.5} & 81.7 &	\textbf{82.1} & \textbf{81.2} & \textbf{72.9} & \textbf{76.8} &	\textbf{76.4} & 73	& \textbf{74.7} & \textbf{82.9} \\
\cdashline{2-15}
& \citet{cattan2021crossdocument} & 85.7 & 81.7 &	83.6 & 70.7 & 74.8 & 72.7 &	59.3 & 67.4 & 63.1 & 56.8 &	65.8& 61	& 73.1 \\\
& \citet{allaway2021sequential} & 83.9 &	84.7 &	84.3 & 74.5 & 70.5 & 72.4 &	70 & 68.1 &	69.2 & - & - & - &	75.3 \\

& Ours &&&&&&&&&&&& \\
&   -- RoBERTa-large & 83.1 & 91.6 & 87.1 & 72.2 & 90.4 & 80.3 &	81.1 &	66.5 &	73.1 &	63.7 &	\textbf{79.3} &	70.6 &	80.2 \\
 & -- RoBERTa-base &  77.2 &	 91.1 &	 83.6 & 61.6 &	 92.8 &	 74.1 &81.0 & 59.4 &	 68.5 & 52.2 &	 79.2 &	 62.9 &	75.4 \\
 & -- BERT-large &  72.8 &	 90.7 &	 80.8 &	 58.1 &	 92.7 &	 71.4 &	 81.8 &	 55.6 &	 66.2 &	 49.0 &	 76.6 &	 59.7 &	72.8  \\
 & -- BERT-base &  69.9 &'	 88.7 &	 78.2 &  55.5 &	 90.9 &	 68.9 & 78.5 & 52.2 &	 62.7 & 45.0 & 72.3 &	 55.5 &	69.9\\
 &   -- RoBERTa-large + TextRank & 75.6 & 91.2 &	82.7 & 59.1  & \textbf{93.2} & 72.3 & 80.8 &	57.4  &	67.1 & 49.6 & 78.9 & 60.9 &	74.1 \\
\hline
\hline
  \end{tabular}}
\caption{\label{tab:subtopic} Detailed results comparing methods utilizing contextual embedding models and their performance on the ECB+ test corpus using \textit{predicted} topic clusters. Note that the systems of \citet{zeng-etal-2020-event, yu2020paired, caciularu2021crossdocument} require significantly more resources than the others ($n^{2}$ versus $n$ transformer computations). Finally, \citet{allaway2021sequential} uses a BERT-base model and a sequential clustering algorithm whereas ours and \citet{cattan2020streamlining} utilize RoBERTa-large models and an agglomerative clustering algorithm.}
\end{table*}

\begin{table*}
\resizebox{\textwidth}{!}{%
    \begin{tabular}{lllccc|ccc|ccc|ccc|c}
\hline
 & & & \multicolumn{3}{c}{MUC}& \multicolumn{3}{c}{$B^{3}$} & \multicolumn{3}{c}{CEAF-e} & \multicolumn{3}{c}{LEA} & CoNLL \\
 \hhline{~~~|---|---|---|---|-}
 & &  &R  & P & F1 &   R   &     P       & F1  &  R &  P       & F1    &   R   &     P     & F1  &   F1 \\
\hline
\multirow{6}{0.75cm}{\centering Events} & \multirow{3}{1cm}{\centering Gold Topics} &  Baseline & 72.9 &  72.4 & 72.7 & 69.7 &	73.5 &	71.5 & 71.1 &	71.7 & 71.4 & 53.5 & 59.2 &	56.1 & 71.9 \\
& &\citet{cattan2021realistic}& 80.1 & 76.3 &	78.1 &	77.4 &	71.7 & 74.5 & 73.1 & 77.8 & 75.4 &	62.9 &	59.1 & 61 & 76 \\
& & Ours & \textbf{87.8} & \textbf{82.9} & \textbf{85.3} & \textbf{86.5} & \textbf{83.1} & \textbf{84.8} & 76.9 & \textbf{82.8} & \textbf{79.7} &	\textbf{74.4} & \textbf{74.0} & \textbf{74.2} & \textbf{83.3} \\
\cline{2-16}
& \multirow{3}{1cm}{\centering Corpus} & Baseline & 72.9	& 60.5 & 66.1 &	69.7 & 56.4 & 62.4 & 51.5 &	68.6 & 58.8 & 45.3 & 42.6 &	43.9 & 62.4 \\
& &\citet{kenyon-dean-etal-2018-resolving}$^{\dagger}$ & 67	& 71 &	69 & 71 & 67 &	69 & \textbf{71} & 67 & 69 & - &	- & - & 69\\
& & Ours &  \textbf{86.4} & \textbf{74.9} & \textbf{80.2} & \textbf{85.3} & \textbf{67.9} & \textbf{75.6} & 65.3 & \textbf{80.1} & \textbf{71.9} &	\textbf{68.3} &	\textbf{57.5} & \textbf{62.4}	& \textbf{75.9} \\
\hline
\hline
\multirow{5}{1cm}{\centering Entities} & \multirow{3}{1cm}{\centering Gold Topics} & Baseline & 61.6 & 85.9 & 71.8 &	48.6 &	\textbf{89} & 62.9 & 76.7 & 45.9 & 57.4 & 37.3 &	65.5 & 47.5	& 64\\
& & \citet{cattan2021crossdocument}& - & - &	- &	- &	-& -& - & - & - &- &- & - & 70.9 \\
& & Ours & \textbf{84.5} & \textbf{90.1} & \textbf{87.2} &	\textbf{79.3} & 86.6 & \textbf{82.8} & \textbf{78.7} & \textbf{68.6} & \textbf{73.3} & \textbf{70.3} & \textbf{75.7} &	\textbf{72.9} &	\textbf{81.1} \\
\cline{2-16}
& \multirow{2}{1cm}{\centering Corpus} & Baseline & 61.9	& 77.5 & 68.8 &	48.7 & 79.6 & 60.4 & 68.2 &	46.1 & 55 & 35.2 & 57.8 & 43.7 & 61.4\\
& & Ours & \textbf{83.9} & \textbf{86.6} & \textbf{85.2} & \textbf{78.5} & \textbf{82.7} & \textbf{80.5} & 73.0 & \textbf{67.9} & \textbf{70.4} & \textbf{67.8} & \textbf{71.7} & \textbf{69.7} & \textbf{78.7}\\
\hline
  \end{tabular}}
\caption{\label{tab:w_singletons} Combined within- and cross-document coreference scores for entities and events \textit{with} singletons, using gold mentions. Gold topics use the ECB+ topics as the initial document pre-clustering while corpus level results do not use any document pre-clustering. We note that the system proposed by \citet{kenyon-dean-etal-2018-resolving} does not use contextual embeddings whereas ours and \citet{cattan2021crossdocument} make use of RoBERTa-large. To the best of our knowledge, we have the only results at the corpus level for entities. \textbf{Bold} values indicate best overall for a particular data subset.}
\end{table*}

\end{document}